\documentclass{article}
\usepackage[a4paper, margin=1in]{geometry}

\pdfoutput=1
\usepackage{times}
\usepackage{graphicx} % more modern
\usepackage{subfigure} 
\usepackage{times}
\usepackage{authblk}

\usepackage{algorithm}
\usepackage{algorithmic}

\usepackage{hyperref}

\usepackage{amsmath,amssymb,amsthm}

\usepackage{natbib}

\usepackage{caption}

\title{\rule{\linewidth}{4pt}\\ \vspace{0.2cm} \Huge \bf Deep Reinforcement Learning Discovers Internal Models\\ \vspace{-0.2cm}\line(1,0){450}}

\author{Nir Baram*}
\author{Tom Zahavy*}
\author{Shie Mannor}
\affil{\{nirb@campus, tomzahavy@campus, shie@ee\}.technion.ac.il\\Electrical Engineering Technion, Israel\\*Equal contribution}
\date{}
\begin{document} 	
\maketitle
\begin{abstract}
Deep Reinforcement Learning (DRL) is a trending field of research, showing great promise in challenging problems such as playing Atari, solving Go and controlling robots. While DRL agents perform well in practice we are still lacking the tools to analayze their performance. In this work we present the Semi-Aggregated MDP (SAMDP) model. A model best suited to describe policies exhibiting both spatial and temporal hierarchies. We describe its advantages for analyzing trained policies over other modeling approaches, and show that under the right state representation, like that of DQN agents, SAMDP can help to identify skills. We detail the automatic process of creating it from recorded trajectories, up to presenting it on t-SNE maps. We explain how to evaluate its fitness and show surprising results indicating high compatibility with the policy at hand. We conclude by showing how using the SAMDP model, an extra performance gain can be squeezed from the agent.
\end{abstract}

\section{Introduction}
Deep Q Network (DQN) is an off-policy learning algorithm that uses a Convolutional Neural Network (CNN; \citep{Krizhevsky2012}) to represent the action-value function. Agents trained using DQN are showing superior performance on a wide range of problems \citep{mnih2015human}. Their success, and that of Deep Neural Network (DNN) in general, is explained by its ability to learn good representations automatically. Unfortunately, its high expressiveness is also the source of its unclarity, making it very hard to analyze. Visualization methods for DNN try to tackle this problem by analyzing and interpreting the learned representations \citep{zeiler2014visualizing,erhan2009visualizing,yosinski2014transferable}. However, these methods were developed for supervised learning tasks, assuming the data is i.i.d, thus overlooking the temporal structure of the learned representation. 

A major challenge in Reinforcement Learning (RL) is scaling to higher dimensions in order to solve real-world applications. Spatial abstractions such as state aggregation \citep{bertsekas1989adaptive}, tries to tackle this problem by grouping states with similar characteristics such as policy behaviour, value function or dynamics. On the other hand, temporal abstractions (i.e., options or skills \citep{Sutton1999}) can help an agent to focus less on lower level details of a task and more on high level planning \citep{dietterich2000hierarchical,parr1998flexible}. The problem with these methods is that finding good abstractions is typically done manually which hampers their wide use.
The internal model principle \citep{francis1975internal}, "Every good key must be a model of the lock it opens",  was formulated mathematically for control systems by \citet{sontag2003adaptation}, claiming that if a system is solving a control task, it must necessarily contain a subsystem which is capable of predicting the dynamics of the system. In this work we follow the same line of thought and claim that DQNs are learning an underlying spatio-temporal model of the problem, without implicitly being trained to. We identify this model as an Semi Aggregated Markov Decision Process (SAMDP), an approximation of the true MDP that allows human interpretability.
\citet{Zahavy2016} used hand-crafted features in order to interpret policies learned by DQN agents. They revealed that DQNs are automatically learning spatio-temporal representations such as hierarchical state aggregation and skills. The main drawback of their approach is that they used a manual reasoning of a t-Distributed Stochastic Neighbour Embedding (t-SNE) map \citep{van2008visualizing}, a tedious process that requires careful inspection as well as an experienced eye. Moreover, their claim to observe skills is not supported with any quantitative evidence. In contrast, we use temporal aware clustering algorithms in order to aggregate the state space, and automatically reveal the underlying spatio-temporal structure of the t-SNE map. The aggregated states uniquely identify skills and allow us to estimate the SAMDP transition probabilities and reward signal empirically. In particular our main contributions are
\begin{enumerate}
\item \textbf{SAMDP:} a model that gives a simple explanation on how DRL agents solve a task - by hierarchically decomposing it into a set of sub-problems and learning specific skills at each.

\item \textbf{Automatic analysis:} we suggest quantitative criteria that allows us to select good models and evaluate their consistency.

\item \textbf{Interpretation:} we developed a novel visualization tool that gives a qualitative understanding of the learned policy.

\item \textbf{Shared autonomy:} the SAMDP model allows us to predict situations where the DQN agent is not performing well. In such occasions we suggest to take the control from the agent and ask for expert advice.
\end{enumerate}    

\section{Background}
We briefly review the standard reinforcement learning framework of discrete-time, finite Markov decision processes (MDPs). In this framework, the goal of an RL agent is to maximize its expected return by learning a policy $\pi:S \rightarrow \Delta_A$, a mapping from states $s \in S$ to probability distribution over actions $A$. At time $t$ the agent observes a state $s_t \in S$, selects an action $a_t \in A$, and receives a reward $r_t$. Following the agents action choice, it transitions to the next state $s_{t+1} \in S$. We consider infinite horizon problems where the cumulative return at time $t$ is given by $R_t = \sum_{t'=t}^\infty \gamma^{t'-t}r_t$, and $\gamma\in[0,1]$ is the discount factor. The action-value function $Q^{\pi}(s,a) = \mathbb{E} [R_t|s_t = s, a_t = a, \pi]$ represents the expected return after observing state $s$, taking action $a$ after which following policy $\pi$. The optimal action-value function obeys a fundamental recursion known as the optimal Bellman Equation: $Q^* (s_t,a_t)=\mathbb{E} \left[r_t+\gamma \underset{a'}{\mathrm{max}}Q^*(s_{t+1},a') \right]
.$\\
\textbf{Deep Q Networks:} The DQN algorithm approximates the optimal Q function using a CNN. The training objective it to minimize the expected TD error of the optimal Bellman Equation: $$\label{DQN_loss}\mathbb{E}_{s_t,a_t,r_t,s_{t+1}}\left\Vert Q_{\theta}\left(s_{t},a_{t}\right)-y_{t}\right\Vert _{2}^{2}$$ \citep{mnih2015human}. DQN is an offline learning algorithm that collects experience tuples $\left\{ s_{t,}a_{t},r_{t},s_{t+1},\gamma\right\}$ and stores them in the \textbf{Experience Replay (ER)} \citep{lin1993reinforcement}. At each training step, a mini-batch of experience tuples are sampled at random from the ER. The DQN maintains two separate Q-networks. The current Q-network with parameters $\theta$, and the target Q-network with parameters $\theta_{target}$. The parameters $\theta_{target}$ are set to $\theta$ every fixed number of iterations. In order to capture the MDP dynamics, the final DQN representation is a concatenation of several consecutive states.

\textbf{Skills, Options, Macro-actions,} \citep{Sutton1999} are temporally extended control structures, denoted by $\sigma$. A skill is defined by a triplet: $\sigma = <I,\pi,\beta>.$ I defines the set of states where the skill can be initiated. $\pi$ is the intra-skill policy, and $\beta$ is the set of termination probabilities determining when a skill will stop executing. $\beta$ is typically either a function of state $s$ or time $t$. Any MDP with a fixed set of skills is a \textbf{Semi-Markov Decision Process (SMDP)}. Planning with skills can be performed by learning for each state the value of choosing each skill. More formally, an SMDP can be defined by a five-tuple $<S, \Sigma, P, R, \gamma>,$ where $S$ is the set of states, $\Sigma$ is the set of skills, $P$ is the SMDP transition matrix, $\gamma$ is the discount factor and the SMDP reward is defined by:
\begin{equation}
\label{eq:skill_reward}
R_s^{\sigma} = \mathbb{E}[r_s^{\sigma}] = \mathbb{E}[r_{t+1} + \gamma r_{t+2} + \cdot\cdot\cdot + \gamma ^{k-1} r_{t+k} | s_t=s,\sigma] 
\end{equation}
The \textbf{Skill Policy} $\mu : S\rightarrow \Delta_\Sigma$  is a mapping from states to a probability distribution over skills. The action-value function $Q_\mu(s, \sigma) = \mathbb{E} [\sum ^\infty _{t=0} \gamma ^t R_t |(s, \sigma), \mu] $ represents the value of choosing skill $\sigma \in \Sigma$ at state $s \in S$, and thereafter selecting skills according to policy $\mu$.
The optimal skill value function is given by: $
\label{OptionBellman}
Q_{\Sigma}^*(s,\sigma) = \mathbb{E} [R_s^{\sigma} + \gamma ^k \underset{\sigma'\in \Sigma}{\mathrm{max}} Q_{\Sigma}^*(s',\sigma')] \enspace$ \citep{stolle2002learning}.

\section{Semi Aggregated Markov Decision Processes}

Reinforcement Learning problems are typically modeled using the MDP formulation. The abundant theory developed for MDP throughout the years gave rise to various algorithms for efficiently solving MDPs, and finding good policies. MDP however, is not the optimal modeling choice when one wishes to analyze a given policy. Policy analysis methods typically suffer from the cardinality of the state space and the length of the planning horizon. For example, building a graphical model that explains the policy will be too large (in terms of states), and complex (in terms of planning horizon) for a human to comprehend.
If the policy one wishes to analyze is known to be planning using temporally-extended actions (i.e. skills), then one may resort to \textbf{SMDP} modeling. The SMDP model reduces the planning horizon dramatically and simplifies the graphical model. There are two problems however with this approach. First, it requires to identify the set of skills used by the policy, a long-standing challenging problem with no easy solution. Second, one is still facing the high complexity of the state space. 

\begin{figure}[h]
\begin{tabular}{ c @{\hskip 0.35in} c }
   \includegraphics[width=0.25\textwidth]{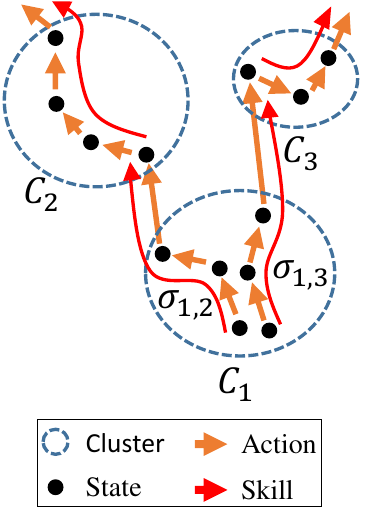} 

 &  \includegraphics[width=0.65\textwidth]{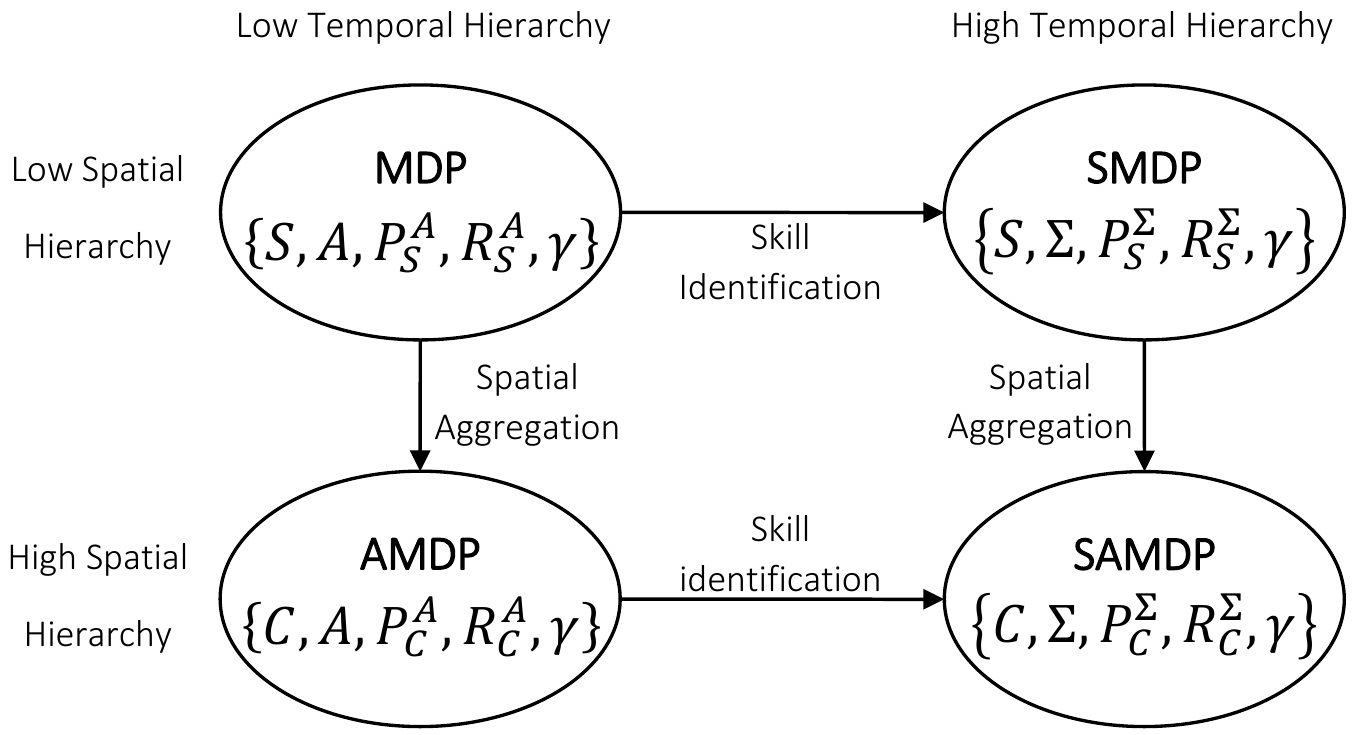} 
\end{tabular}
\caption{\textbf{Left:} Illustration of state aggregation and skills. Primitive actions (orange arrows) cause transitions between MDP states (black dots) while skills (red arrows) induce transitions between SAMDP states (blue circles). \textbf{Right:} Modeling approaches for analyzing policies. MDP (top-left): a policy is analyzed in the MDP state space $S$, with the original set of primitive actions $A$. SMDP (top-right): using the set of identified skills $(A \rightarrow \Sigma)$, the policy is easier to analyze. AMDP (bottom-left): State aggregation allows to reduce state space complexity $(S \rightarrow C)$. SAMDP  (bottom-right): identifying skills in the AMDP model reduces the planning horizon $(S \rightarrow C ,A \rightarrow \Sigma)$.}
\label{fig:modeling}
\end{figure}

A different modeling approach is to aggregate similar states first. This is useful when there is a reason to believe that groups of states share common attributes such as similar policy, value function or dynamics. State aggregation is a well studied problem that can be solved by applying clustering on the MDP state representation.  
These models are not necessarily Markovian, however they can provide great simplification of the state space. With a slight abuse of notation we denote this model as Aggregated MDP (\textbf{AMDP}). Under the right state-representation, the AMDP can also help to identify skills (if exist). We argue that this is possible if the AMDP dynamics is such that the majority of the transitions are done within the clusters, followed by rare transitions between clusters. As we will show in the experiments section, DQN indeed provides a good state representation that allows skill identification.

If the state representation contains both spatial and temporal hierarchies, then the AMDP model can be further simplified into an \textbf{SAMDP} model. Under SAMDP modeling, both the state-space cardinality and the planning horizon are reduced, making policy reasoning more feasible. We summarize our observations about the different modeling approaches in Figure~\ref{fig:modeling}.\\
In the remaining of this section we explain the SAMDP modeling in detail and focus on explaining how to empirically build an SAMDP model from experience. To do so we explain how to aggregate states, identify skills and estimate the transition probabilities and reward measures. Finally we discuss how to evaluate the fitness of an empiric SAMDP model to the data.

\subsection{State aggregation}
\label{sec:agg}
We evaluate a \textbf{DQN} agent, by letting it play multiple trajectories with an $\epsilon$-greedy policy. During evaluation we record all visited states, neural activations, value estimations, and index them by their visitation order. We treat the neural activations as the state representation that the DQN agent has learned. \citet{Zahavy2016} showed that this state representation captures a spatio-temporal hierarchy and therefore makes a good candidate for state aggregation. We then apply \textbf{t-SNE} on the neural activations data, a non-linear dimensionality reduction method that is particularly good at creating a single map that reveals structure at many different scales. t-SNE reduces the tendency of points to crowd together in the center of the map by using a heavy tailed Student-t distribution in the low dimensional space. The result is a compact, well separated representation, that is easy to visualize and interpret.

We represent an MDP state $s_i$ by a feature vector $x_i\in\mathbb{R}^3$, comprised of the two t-SNE coordinates and the DQN value estimate. Using this representation we aggregate the state space by applying clustering algorithms and define the AMDP \textbf{states} $C$ as the resulting \textbf{clusters}. Standard clustering algorithms assume that the data is drawn from an i.i.d distribution, however our data is generated from an MDP which violates this assumption.

\begin{algorithm}
\caption{K-means \citep{macqueen1967some} for state aggregation}          % give the algorithm a caption
\label{alg:kmeans} 
\textbf{Input:} MDP sates feature representation $(x_1, x_2, \cdots, x_n).$\\
\textbf{Output:} SAMDP states $(c_1, c_2, \cdots, c_k).$\\
\textbf{Objective:} minimize the within-cluster sum of squares:
\begin{center}
$\underset{\mathbf{C}} {\operatorname{arg\,min}}  \sum_{i=1}^{k} \sum_{\mathbf x \in C_i} \left\| \mathbf x - \boldsymbol\mu_i \right\|^2$  \end{center}
where $\mu_i$ is the mean of points in $c_i$.\\
\textbf{Repeat} until convergence:
\begin{enumerate}
\item \textbf{Assignment step,} each observation $x_i$ is assigned to its closest cluster center:
\begin{center}
$ C_i^{(t)} = \big \{ x_p : \big \| x_p - \mu^{(t)}_i \big \|^2 \le \big \| x_p - \mu^{(t)}_j \big \|^2 \forall j, 1 \le j \le k \big\}. $
\end{center}
\item \textbf{Update step,} each cluster center $\mu_j$ is updated to be the mean of its constituent instances:
\begin{center} $\mu^{(t+1)}_i = \frac{1}{|C^{(t)}_i|} \sum_{x_j \in C^{(t)}_i} x_j.$ \end{center}
\end{enumerate}
\end{algorithm}

In order to alleviate this problem, we suggest two versions of K-Means (Algorithm~\ref{alg:kmeans}) that take into account the temporal structure of the data. \textbf{(1) Spatio-Temporal Cluster Assignment} that encourages temporal coherency by modifying the assignment step in the following way:
\begin{equation}
\label{convolve}
C_i^{(t)} = \big \{ x_p : \big \| X_{p-w:p+w} - \mu^{(t)}_i \big \|^2 \le \big \| X_{p-w:p+w} - \mu^{(t)}_j \big \|^2, \forall j, 1 \le j \le k \big\}
\end{equation}
Where $p$ is the time index of observation $x_p$, $X_{p-w:p+w}$ is the set of $2w$ points before and after $x_p$ along the trajectory. In this way, a point $x_p$ is assigned to a cluster $\mu_j$, if its neighbours along the trajectory are also close to $\mu_j$.\\
\textbf{(2) Entropy Regularization Cluster Assignment} that creates simpler models by adding an entropy regularization term to the K-mean assignment step:
\begin{equation}
\label{entropy_regularization}
C_i^{(t)} = \big \{ x_p : \big \| x_p - \mu^{(t)}_i \big \|^2 + d \cdot e^{t-1}_{x_p \rightarrow i} \le \big \| x_p - \mu^{(t)}_j \big \|^2   + d \cdot e^{t-1}_{x_p \rightarrow j}, \forall j, 1 \le j \le k \big\}.
\end{equation}
Where $d$ is a penalty weight, and $e^{t-1}_{x_p \rightarrow i}$ indicates the entropy (as defined in Section~\ref{sub:eval}) gain of changing $x_p$ assignment to cluster $i$ in the SMDP obtained at iteration $t-1$. This is equivalent to minimizing an energy function, the sum of the K-means objective function and an entropy term.\\
We also considered \textbf{Agglomerative Clustering,} a bottom-up hierarchical approach. Starting with a mapping from points to clusters (e.g., each point is a singular cluster), the algorithm advances by merging pairs of clusters such that a linkage criteria is minimized. In order to encourage temporal coherency in cluster assignments we define a new linkage criteria based on \citet{Inchoate:Ward63}:
\begin{equation}
c(A,B)=(1-\lambda) \cdot mean\{\|x_a-x_b\|:a \in A, b \in B\}+ \lambda\cdot e_{\{A,B\} \rightarrow AB}
\end{equation}
where $e_{\{A,B\} \rightarrow AB}$ measures the difference between the entropy of the corresponding SMDP before and after merging clusters $A,B$.
\subsection{Temporal abstractions}
We define the SAMDP \textbf{skills} by their initiation and termination AMDP states $C$: 
\begin{equation}
\label{eq:skill_def}
\sigma_{ij} = <\{ c_i \},\pi_{i,j},\{ c_j \}>.
\end{equation}

More implicit, once the DQN agent enters an AMDP state $c_i$ at an MDP state $s_t \in c_i$, it follows the skill policy $\pi_{i,j}$ for $k$ steps, until it reaches a state $s_{t+k} \in c_j$, s.t $i \neq j$. Note that we do not define the skill policy implicitly, but we will observe later that our model successfully captures spatio-temporal defined skill policies. We set the SAMDP discount factor $\gamma$ same as was used to train the DQN. We now turn to estimate the SAMDP probability matrix and reward signal. For that goal we make the following assumptions:
 
\textbf{Definition 1. } \textit{ A deterministic probability matrix, is a probability matrix such that each of its rows contains one element that equals to $1$ and the others equal to $0$.}

%\begin{}
\textbf{Assumption 1. } \textit{The MDP transition matrices $P_A: P^{a \in A}_{i,j}=Pr(x_j|x_i,a)$ are deterministic.}\\
%\end{assumption}
This assumption limits our analysis for environments with deterministic dynamics. However, many interesting problems are in fact deterministic, e.g., Atari2600 benchmarks, Go, Chess etc.

%\begin{assumption}
\textbf{Assumption 2. } \textit{
The policy played by the DQN agent is deterministic.}\\
Although DQN chooses actions deterministically (by selecting the action that corresponds to the maximal Q value in each state), we allow $5\%$ $\epsilon$ stochastic exploration. This introduces errors into our model that we will later analyze.\\
Given the DQN policy, the MDP is reduced into a Markov Reward Process (MRP) with probability matrix $P^{\pi^{DQN}}_{i,j}=Pr(x_j|x_i,a=\pi^{DQN}(x_i))$. Note that by Assumptions 1 and 2, this is also a deterministic probability matrix. 

The SAMDP \textbf{transition probability matrix} $P_\Sigma: P^{\sigma \in \Sigma}_{i,j}=Pr(c_j|c_i,\sigma)$, indicates the probability of moving from state $c_i$ to $c_j$ given that skill $\sigma$ is chosen. It is also a deterministic probability matrix by our definition of skills (Equation~\ref{eq:skill_def}). Our goal is to estimate the probability matrix that the DQN policy induces on the SAMDP model: $P^{\pi^{DQN}}_{i,j}=Pr(c_j|c_i,\sigma=\pi^{DQN}(c_i))$. 

We do not require this policy to be deterministic from two reasons. First, we evaluate the DQN agent with an $\epsilon$-greedy policy. While almost deterministic in the view of a single time step, the variance of its behaviour increases as more moves are played. Second, the aggregation process is only an approximation. For example, a given state may contain more than one "real" state and therefore hold more than one skill with different transitions. A stochastic policy can solve this disagreement by allowing to choose skills at random. 

This type of modeling does not guarantee that our SAMDP model is Markovian and we are not claiming it to be. SAMDP is an approximation of the the true dynamics that simplifies it over space and time to and allow human interpretation. Finally, we estimate the skill length $k_\sigma$ and SAMDP reward for each skill from the data using Equation~\ref{eq:skill_reward}. In the experiments section we show that this model is in fact consistent with the data by evaluating its value function:
\begin{equation}
\label{eq:samdp_value}
V_{SAMDP} = ( I+\gamma^{k}P )^{-1}r
\end{equation}
and the greedy policy with respect to it:
\begin{equation}
\label{eq:samdp_greedy_policy}
\pi_{greedy}(c_i) = \underset{j}{\mbox{argmax}} \{ R_{\sigma_{i,j}}+\gamma^{k_{\sigma_{i,j}}}v_{SAMDP}(c_j) \}
\end{equation}

\subsection{Evaluation criteria}
\label{sub:eval}
We follow the analysis of \citep{hallak2013model} and define criteria to measure the fitness of a model empirically. We define the \textbf{Value Mean Square Error(VMSE)} as the normalized distance between two value estimations: $\mbox{VMSE} = \frac{\| v^{DQN}-v^{SAMDP} \|}{\|v^{DQN}\|}.$ The SAMDP value is given by Equation~\ref{eq:samdp_value} and the DQN value is evaluated by averaging the DQN value estimates over all MDP states in a given cluster (SAMDP state): ${v^{DQN}(c_j)}=\frac{1}{|C_j|}\sum_{i: s_i \in c_j}v^{DQN}(s_i)$ .\\
The \textbf{Minimum Description Length} (MDL; \citep{rissanen1978modeling}) principle is a formalization of the celebrated Occam’s Razor. It copes with the over-fitting problem for the purpose of model selection. According to this principle, the best hypothesis for a given data set is the one that leads to the best compression of the data. Here, the goal is to find a model that explains the data well, but is also simple in terms of the number of parameters. In our work we follow a similar logic and look for a model that best fits the data but is still “simple”.\\
Instead of considering "simple" in terms of the number of parameters, we measure the simplicity of the spatio-temporal state aggregation. For spatial simplicity we define the Inertia: $I = \sum_{i=0}^{n}\min_{\mu_j \in C}(||x_j - \mu_i||^2)$ which measures the variance of MDP states inside a cluster (AMDP state). For temporal simplicity we define the entropy: $e= - \sum_i \{ |C_i| \cdot \sum_j{P_{i,j} \log P_{i,j}} \}$ , and the \textit{Intensity Factor} which measures the fraction of in/out cluster transitions: $F = \sum_j \frac{P_{jj}}{\sum_i P_{ji}}.$\\
To \textbf{summarize}, the stages of building an SAMDP model are:
\begin{enumerate}
\item \textbf{Evaluate :} Run the trained (DQN) agent, record visited states, representations and Q-values.
\item \textbf{Reduce :} Apply t-SNE on the state representations to obtain a low dimensional map.
\item \textbf{Aggregate :} Cluster states in the map.
\item \textbf{Model :} Fit an SAMDP model, select the best model. 
\item \textbf{Visualize :} Visualize the SAMDP on top of the t-SNE map.
\end{enumerate}

\section{Experiments}
\label{Experiments}
\textbf{Setup.} We evaluate our method on three Atari2600 games, Breakout, Pacman and Seaquest. For each game we collect 120k game states (each represented by 512 features), and Q-values for all actions. We apply PCA to reduce the data to 50 dimensions, then we apply t-SNE using the Barnes Hut approximation to reach the desired low $2$ dimension. We run the t-SNE algorithm for 3000 iterations with perplexity of 30. We use Spatio-Temporal K-means clustering (Section~\ref{sec:agg}) to create the AMDP states (clusters), and evaluate the transition  probabilities between them using the trajectory data. We overlook flicker-transitions where a cluster is visited for less than $f$ time steps before transiting out. Finally we truncate transitions with less than 0.1 probability.
%\begin{wrapfigure}{r}{\textwidth}
%\centering
%\captionsetup{justification=centering}
%\vspace{-2pt}
%\caption{\textbf{Model Selection:} Correlation between criteria pairs for the SAMDP model of Breakout.}
%\label{fig:model_sel}
%\end{wrapfigure}

%
%\begin{wrapfigure}[16]{r}{0.6\textwidth}
%\vspace{-15pt}
%\centering
%\vspace{-6pt}
%\caption{\textbf{Model Evaluation.} \textbf{Top:} Value function consistency. \textbf{Center:} greedy policy correlation with trajectory reward. \textbf{Bottom:} top (blue), least (red) rewarded trajectories.}
%\label{fig:model_consis}
%\end{wrapfigure}

\begin{figure}[h]
\centering
\captionsetup{justification=centering}
\includegraphics[trim=1cm 0.5cm 1cm 1cm,clip,width=0.8\textwidth]{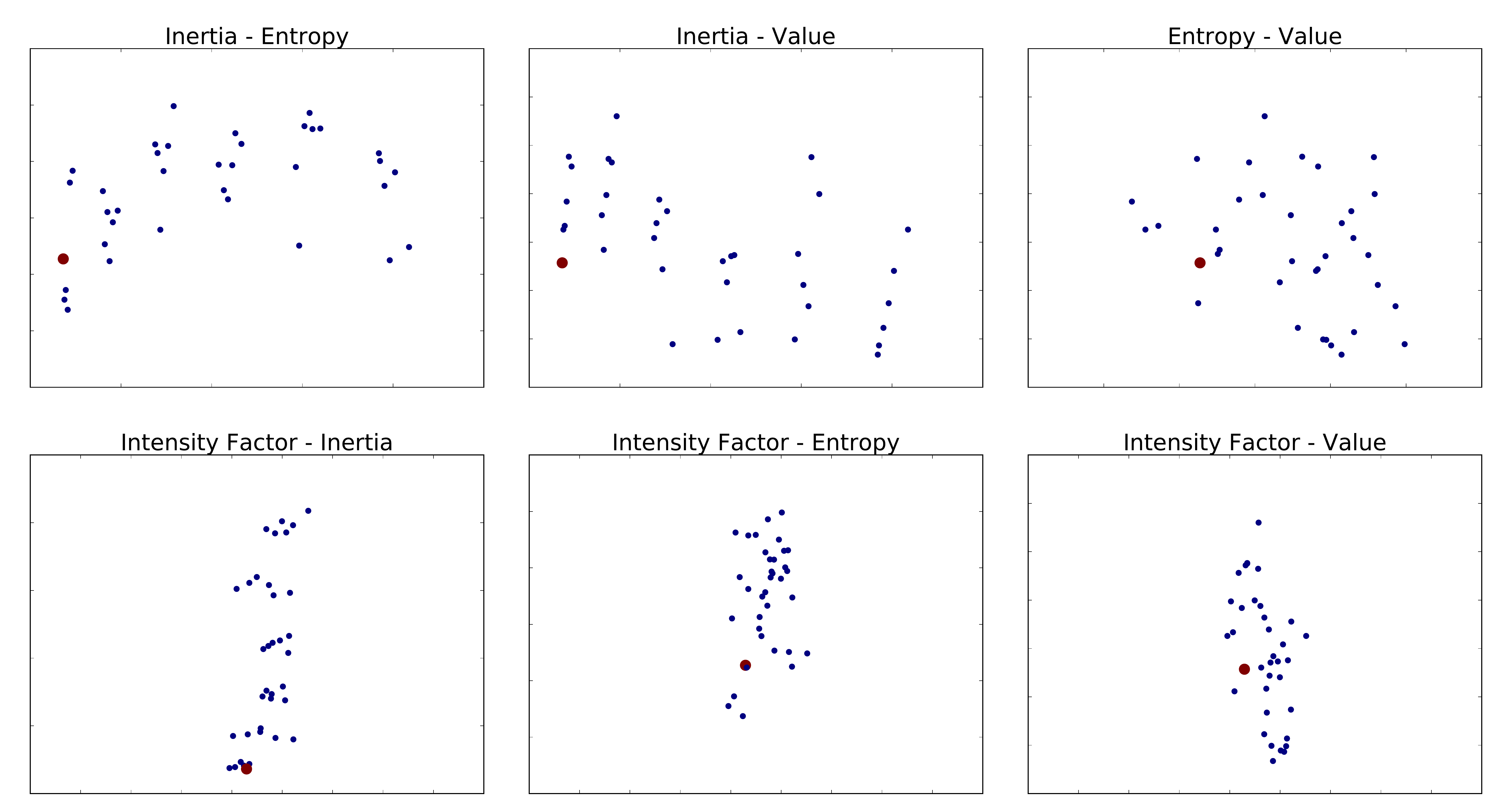}
\vspace{-2pt}
\caption{\textbf{Model Selection:} Correlation between criteria pairs for the SAMDP model of Breakout.}
\label{fig:model_sel}
\end{figure}

\textbf{Model Selection.} We perform a grid search on two parameters: \textit{i}) number of clusters \textbf{$N^c \in [15,25]$}. \textit{ii}) window size \textbf{$w \in [1,7]$}. We found that models larger (smaller) than that are too cumbersome (simplistic) to analyze. We select the best model in the following way: Let $e(w,n),i(w,n),v(e,n),f(e,n)$ be the entropy, inertia, VMSE, and intensity factor respective measures of configuration $(w,n)$ in the greed search. Let
$E=\{e(w,n)\}, I=\{i(w,n)\}, V=\{v(w,n)\}, F=\{f(w,n)\}$ be the corresponding sets grouped over all grid search configurations. We sort each set from good to bad, i.e. from minimum to maximum (except for intensity factor where larger values are considered better). We then iteratively intersect the p-prefix of all sets (i.e. the first p elements of each set) starting with 1-prefix. We stop when the intersection is non empty and choose the configuration at the intersection. Figure~\ref{fig:model_sel} shows the correlation between pairs of criteria (for Breakout).

\begin{figure}[h]
\begin{center}
\includegraphics[width=\textwidth]{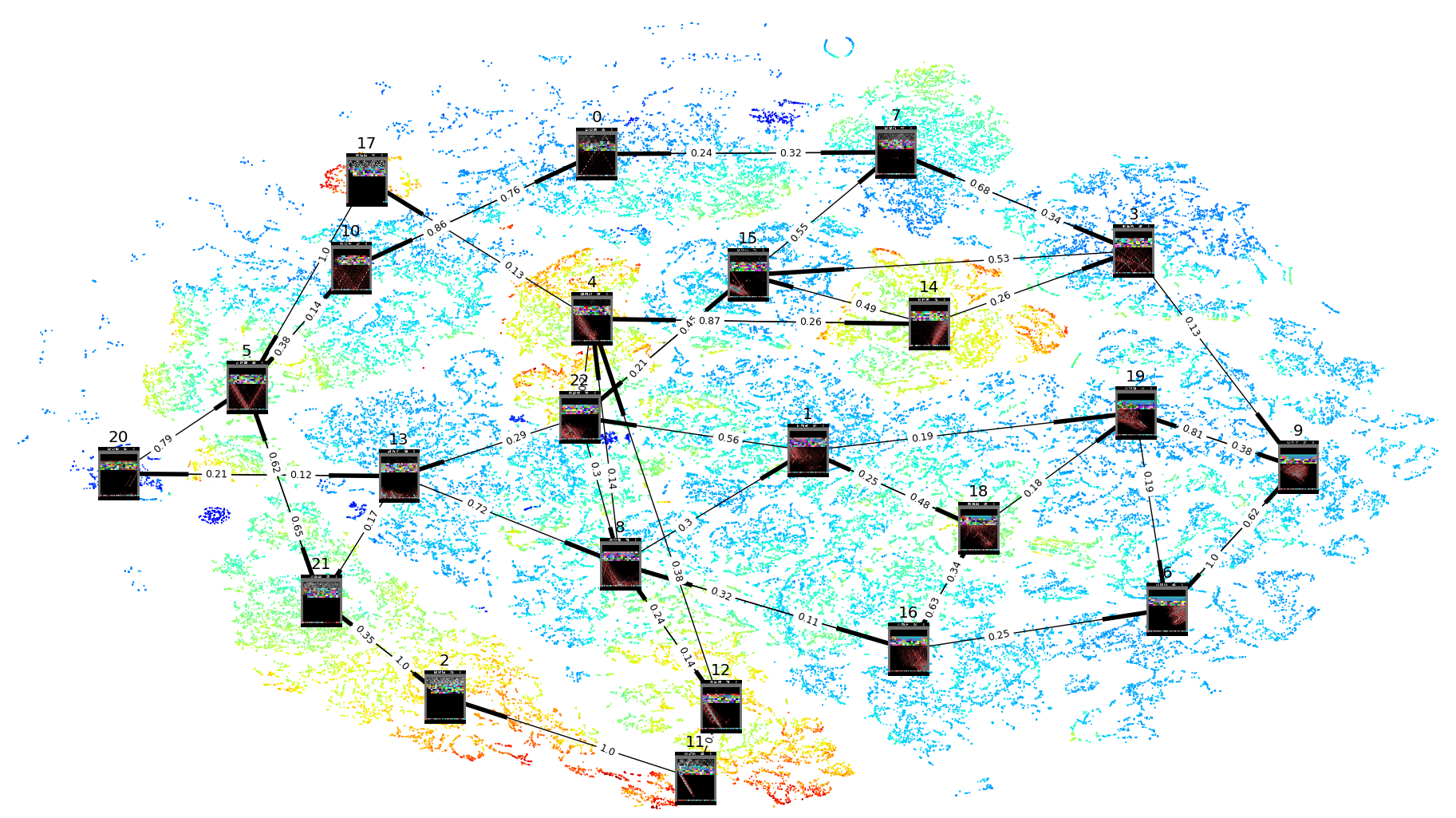} 
\caption{SAMDP visualization for Breakout over the t-SNE map colored by value estimates (low values in blue and high in red).}
\label{fig:Breakout}
\vspace{-5pt}
\end{center}
\end{figure}

Overall, we see a tradeoff between spatial and temporal complexity. For example, in the bottom left plot, we observe correlation between the Inertia and the Intensity Factor; a small window size $w$ leads to well-defined clusters in space (low Inertia) at the expense of a complex transition matrix (small intensity factor). A large $w$ causes the clusters to be more spread in space (large Inertia), but has the positive effect of intensifying the in-cluster transitions (high intensity factor). We also measure the p-value of the chosen model with the null hypothesis being the SAMDP model constructed with randomly clustered states. We tested 10000 random SAMDP models, none of which scored better than the chosen model (for any of the evaluation criteria).\\
\textbf{Qualitative Evaluation.} Examining the resulting SAMDP (Figure~\ref{fig:Breakout}) it is interesting to note the sparsity of transitions. This indicate that clusters are well located in time. Inspecting the mean image of each cluster also reveal some insights about the nature of the skills hiding within. We also see evidence for the "tunnel-digging" option described in \citep{Zahavy2016} in the transitions between clusters 11,12,14 and 4.

\begin{figure}
\centering
\includegraphics[trim=2cm 0cm 0cm 1cm,clip,width=0.8\textwidth]{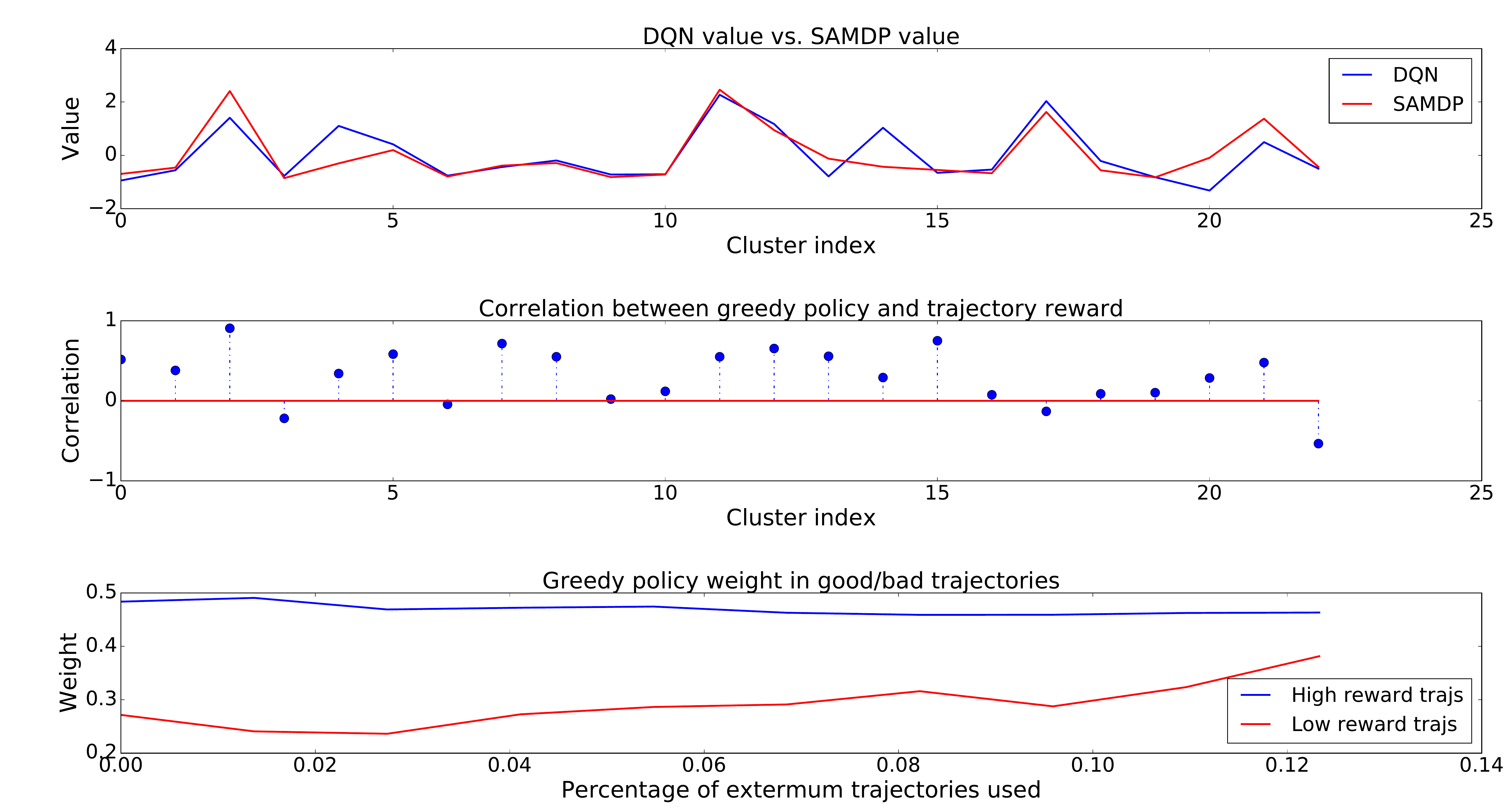}
\vspace{6pt}
\caption{\textbf{Model Evaluation.} \textbf{Top:} Value function consistency. \textbf{Center:} greedy policy correlation with trajectory reward. \textbf{Bottom:} top (blue), least (red) rewarded trajectories.}
\vspace{-4pt}
\label{fig:model_consis}
\end{figure}

\textbf{Model Evaluation.} We evaluate our model using three different methods. First, the VMSE criteria (Figure~\ref{fig:model_consis}, top): high correlation between the DQN values and the SAMDP values gives a clear indication to the fitness of the model to the data. Second, we evaluate the correlation between the transitions induced by the policy improvement step and the trajectory reward $R^j$. To do so, we measure $P_i^j:$ the empirical distribution of choosing the greedy policy at state $c_i$ in that trajectory. Finally we present the correlation coefficients at each state: $corr_i = corr(P_i^j,R^j)$ (Figure~\ref{fig:model_consis}, center). Positive correlation indicates that following the greedy policy leads to high reward. Indeed for most of the states we observe positive correlation, supporting the consistency of the model. The third evaluation is close in spirit to the second one. We create two transition matrices $T^+,T^-$ using k top-rewarded trajectories and k least-rewarded trajectories respectively. We measure the correlation of the greedy policy $T^G$ with each of the transition matrices for different values of k (Figure~\ref{fig:model_consis} bottom). As clearly seen, the correlation of the greedy policy and the top trajectories is higher than the correlation with the bad trajectories.\\
\textbf{Eject Button: Performance improvement.} In the following experiment we show how the SAMDP model can help to improve the performance of a trained policy. The motivation for this experiment stems from the idea of shared autonomy \citep{icra11a}. There are domains where errors are not permitted and performance must be as high as possible. The idea of shared autonomy is to allow an operator to intervene in the decision loop in critical times. For example, it is known that in 20$\%$ of commercial flights, the auto-pilot returns the control to the human pilots. For this experiment we first build an SAMDP model and then let the agent to play new (unseen) trajectories. We project the online state visitations onto our model and monitor its transitions along it. We define $T^+,T^-$ as above. If the likelihood of $T^-$ with respect to the online trajectory is greater than the likelihood of $T^+$, we press the \textbf{Eject} button and terminate this execution (a procedure inspired by option interruption \citep{sutton1999between}). We're interested to measure the average performance of the un-terminated trajectories with respect to all trajectories. The performance improvement achieved with and without using the Eject button is presented in Table~\ref{table:eject}.
\begin{table}[h]
\begin{tabular}{| l | c | c | c |}
\hline
Game & Average Score without eject & Average Score with eject & Improvement $\%$ \\ \hline
Breakout & 293 & \textbf{400} & +36 \\ \hline
Seaquest & 5641 & \textbf{6780} & +20 \\ \hline
Pacman & 230 & \textbf{241} & +4.7 \\ \hline
\end{tabular}
\caption{\textbf{Performance gain using eject button} averaged over 60 trajectories. Numbers are reported for DQN agents we train ourselves.}
\label{table:eject}
\end{table}
\section{Discussion}
\label{Discussion}
In this work we considered the problem of automatically building an SAMDP model for analyzing trained policies. Starting from a t-SNE map of neural activations, and ending up with a compact model that gives a clear interpretation for complex RL tasks. We showed how SAMDP can help in identifying skills that are well defined in terms of initiation and termination sets. However, the SAMDP doesn't offer much information about the skill policy and we suggest to further investigate it in future work. It would also be interesting to see whether skills of different states actually represent the same behaviour. Most importantly, the skills we find are determined by the state aggregation. Therefore, they are impaired by the artifacts of the clustering method used. In future work we will consider other clustering methods that better relate to the topology (such as spectral-clustering), to see if they lead to better skills.

In the Eject experiment we showed how SAMDP model can help to improve the policy at hand without the need to re-train it. It would be even more interesting to use the SAMDP model to improve the training phase itself. The strength of SAMDP in identifying spatio and temporal hierarchies could be used for harnessing DRL hierarchical algorithms \citep{tessler2016deep,kulkarni2016hierarchical}. For example by automatically detecting sub-goals or skills.

Another question we're interested in answering is whether a global control structure exists? Motivated by the success of policy distillation ideas \citep{rusu2015policy}, it would be interesting to see how well an SAMDP built for game A, explains game B?
Finally we would like to use this model to interpret other DRL agents that are not specifically trained to approximate value such as deep policy gradient methods.

\newpage
\footnotesize
\bibliographystyle{plainnat}

\bibliography{Draft}
\end{document}